\documentclass[10pt,twocolumn,letterpaper]{article}

\usepackage[pagenumbers]{cvpr} %
\usepackage{booktabs}   %
\usepackage{tabularx}   %
\usepackage{array}      %
\usepackage{algorithm}
\usepackage{booktabs}
\usepackage{multirow}
\usepackage[noend]{algpseudocode}

\usepackage{cuted} %

\usepackage{currfile} %

\usepackage{caption} %

\definecolor{cvprblue}{rgb}{0.21,0.49,0.74}
\usepackage[pagebackref,breaklinks,colorlinks,allcolors=cvprblue]{hyperref}
\usepackage{capt-of}   
\input{supp_preamble}

\def\confName{CVPR}
\def\confYear{2027}

\title{EvoMisty: Building, Optimizing, and Evolving Capabilities on a Social Robot}

\author{
Xiao Wang \quad Lu Dong \quad Ifeoma Nwogu \quad Srirangaraj Setlur \quad Venu Govindaraju\\
State University of New York at Buffalo\\
{\tt\small \{xwang277, ludong, inwogu, setlur, govind\}@buffalo.edu}
}

\begin{document}
\twocolumn[{%
\renewcommand\twocolumn[1][]{#1}%
\maketitle
\begin{center}
  \vspace{-1.5em}
  \includegraphics[width=\linewidth]{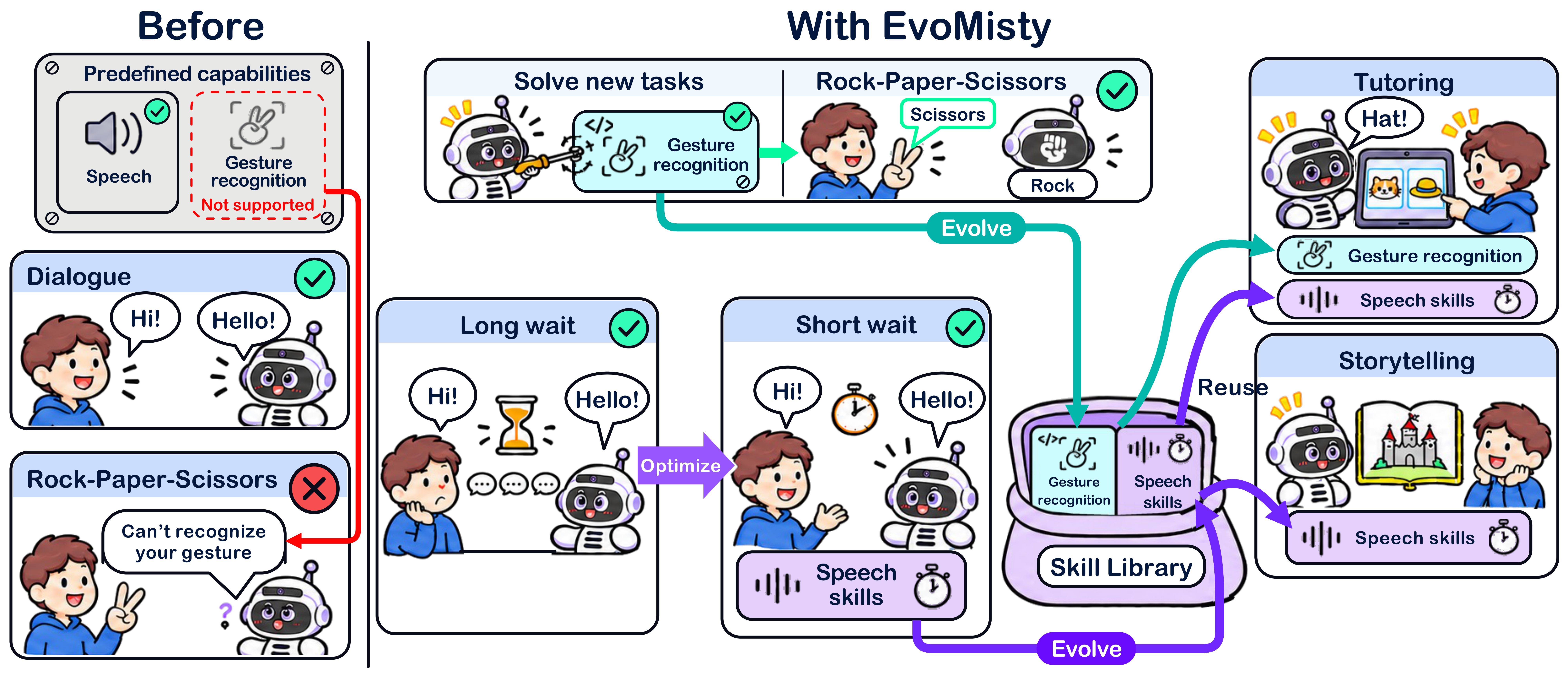}
  \captionof{figure}{\textbf{Before:} Misty's capabilities are predefined,
  so tasks that need a missing capability cannot be completed.
  \textbf{With EvoMisty:} The agent autonomously builds required capabilities, optimizes its solutions on user-specified metrics, and distills these solutions into reusable skills.}
  \label{fig:teaser}
\end{center}%
}]

\begin{abstract}
Social robots face diverse and changing tasks in companionship and education. LLM-based methods can generate task programs, but they still rely on manually developed control interfaces and predefined capabilities. They also lack a general mechanism to optimize completed tasks based on user-specified metrics and reuse performance improvements across tasks.
We present EvoMisty, a plugin-based framework that enables coding agents to build and continually improve robot capabilities. EvoMisty builds a control program from official documentation and verifies candidate capabilities on a physical robot. Verified capabilities are stored in an atomic capability library and combined with existing skills to solve tasks. After task completion, EvoMisty can collect execution data, measure user-specified metrics, and iteratively improve task programs. It then distills reusable skills from final solutions into a skill library for future tasks.
We evaluate EvoMisty on a physical Misty II robot using Claude Code and Codex. The two agents construct 109 and 107 verified atomic capabilities, respectively. Both complete all 28 AutoMisty benchmark tasks and seven additional tasks beyond AutoMisty's capabilities. Optimization improves all seven evaluated metrics across five tasks. For example, dialogue response latency decreases by 52\% and 71\%, respectively. Reusing skills from one optimized dialogue task reduces latency by 33\% to 72\% in three other dialogue tasks without further optimization. \iftoggle{cvprfinal}{\textbf{All code and experimental videos will be publicly released through the webpage: \href{https://wangxiaoshawn.github.io/projects/evomisty/index.html}{EvoMisty}.}}{}
\end{abstract}

\section{Introduction}
\label{sec:intro}

Social robots face diverse and changing tasks in companionship and education. New tasks may require additional device functions, external hardware, or better execution performance. LLM-based methods can generate task programs, but they rely on predefined device capabilities and lack a general mechanism to improve performance.

Existing methods, such as Code as Policies~\cite{liang2023code}, AutoMisty~\cite{wang2025automisty}, and MistyPilot~\cite{wang2026mistypilot}, use predefined interfaces or skills to complete tasks. Recent studies, including ASPIRE~\cite{lu2026aspire} and RoboFoundry~\cite{liang2026robofoundry}, further explore program improvement through execution feedback. However, two challenges remain. First, constructing and verifying basic device capabilities still largely requires manual development. Second, existing approaches mainly focus on task completion and failure correction. They lack a general mechanism to optimize completed tasks based on user-specified performance metrics and transfer these improvements to future tasks.

We introduce EvoMisty, a plugin-based framework that enables coding agents to autonomously build and continually improve robot capabilities. As shown in Fig.~\ref{fig:overview}, EvoMisty consists of four stages. \textbf{Build} creates a control program from official documentation and stores verified capabilities in an atomic capability library. \textbf{Solve} combines atomic capabilities and existing skills to write and test task programs. \textbf{Optimize} collects execution data, measures user-specified metrics, and iteratively improves completed programs. \textbf{Evolve} distills reusable skills from final solutions into a skill library for future tasks.

We evaluate EvoMisty on a physical Misty II robot using Claude Code and Codex. The two agents construct 109 and 107 verified atomic capabilities, respectively. Both complete all 28 AutoMisty benchmark tasks and seven additional tasks beyond AutoMisty's capabilities. Optimization improves all seven evaluated metrics across five tasks. Reusing skills from one optimized dialogue task reduces latency by 33\% to 72\% in three other dialogue tasks without further optimization.

\noindent Our main contributions are as follows:
\begin{itemize}
    \item We propose EvoMisty, a plugin-based framework that enables coding agents to autonomously construct and verify robot capabilities from official documentation.

    \item We introduce a metric-driven optimization workflow that improves completed tasks and transfers performance improvements through reusable skills.

    \item We demonstrate the effectiveness of EvoMisty on a physical social robot using two coding agents, 35 tasks, seven performance metrics, and cross-task reuse experiments.
\end{itemize}

\section{Related Work}
\label{sec:related_work}
\begin{figure*}[t]
    \centering
    \includegraphics[width=\textwidth]{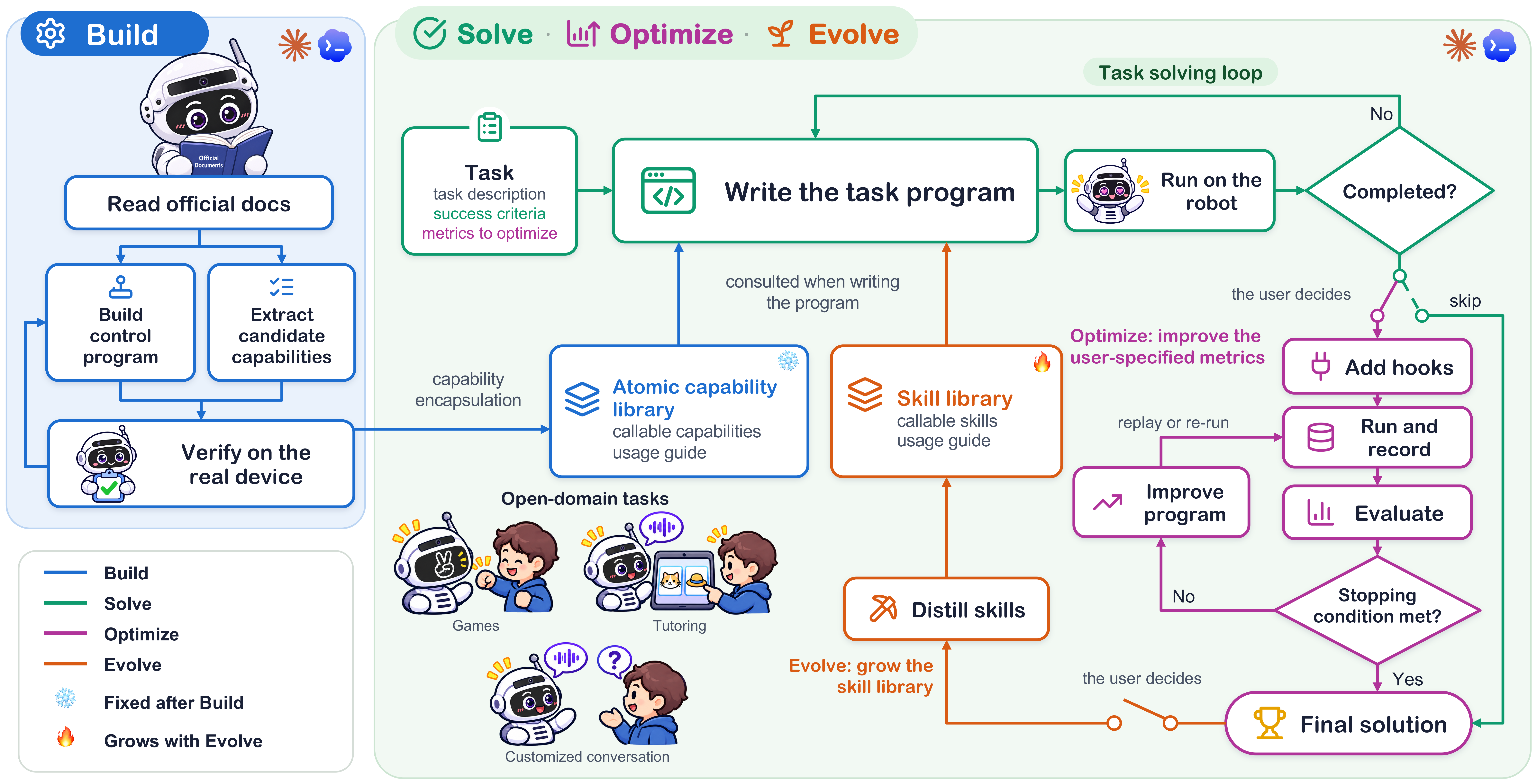}
    \caption{Overview of EvoMisty, consisting of four stages: \textbf{Build}, \textbf{Solve}, \textbf{Optimize}, and \textbf{Evolve}.}

    \label{fig:overview}

    \label{fig:overview}
\end{figure*}

\subsection{LLM-Based Task Solving for Social Robots}
LLM-based methods use program generation and skill orchestration to complete natural-language robot tasks~\cite{raptis2025agentic,li2026roboclaw}.
Code as Policies~\cite{liang2023code} generates executable programs using predefined perception and control interfaces.
Recent coding agents extend this formulation with multi-turn execution feedback~\cite{liu2026rapid,gao2026self,yoon2025moduloop,ahn2026towards}.
CaP-X~\cite{fu2026cap} benchmarks such agents on manipulation and finds that success degrades once human-designed abstractions are removed.
FAEA~\cite{tsui2026demonstration} applies an unmodified software-engineering agent to simulated manipulation, where the agent discovers the simulator APIs by reading their documentation.
URAI~\cite{ge2026make} pairs an agent that writes reusable robot tools with an agent that calls them during execution.
AutoMisty~\cite{wang2025automisty} uses multiple agents to generate and debug robot programs through execution and human feedback.
MistyPilot~\cite{wang2026mistypilot} dynamically selects and orchestrates existing skills for sensor-triggered actions and multi-turn interactions.
LLMs are also reshaping human-robot interaction more broadly~\cite{wang2026we,pinto2025designing}.
On Pepper, a function-calling framework lets the LLM plan over a predefined tool library~\cite{studerus2026framework}.
RoboCritics~\cite{kim2026robocritics} checks LLM-generated robot programs with expert-informed critics over execution traces, and Robo-Blocks~\cite{sato2026robo} scaffolds novice programming of social robots.
SkillComposer~\cite{woods2026skillcomposer} compresses successful programs into reusable macro skills in simulation.
These methods improve task flexibility but largely depend on predefined device interfaces or skills.
In contrast, EvoMisty builds a control program from official documentation, verifies candidate capabilities on a physical robot, and stores verified capabilities in an atomic capability library for future tasks.

\subsection{Agentic Self-Improvement for Robotic Systems}
Recent studies explore autonomous robot improvement through execution feedback~\cite{luo2026toward,zhou2026improves,chen2026gap,elmaaroufi2026rho,wang2026harnesspai,hu2026simex}.
More generally, self-evolving agents update their memory, skills, or tools from past interactions instead of staying fixed after deployment~\cite{gao2025survey,zhou2026self,zhang2026darwin}.
ASPIRE~\cite{lu2026aspire} uses multimodal execution traces to detect failures, repair programs, and accumulate reusable skills.
RoboFoundry~\cite{liang2026robofoundry} extends improvement to supporting agent systems, including context management and skill organization.
ENPIRE~\cite{xiao2026enpire} establishes a physical robot experimentation loop through automated resets, execution, and outcome verification.
RATs~\cite{zhang2026playful} builds a code skill library through self-directed play before downstream tasks arrive.
COAP~\cite{hu2026embodied} has coding agents develop a shared library of stateful control code in a closed loop.
Kuroki et al.~\cite{kuroki2026learning} archive closed-loop programs improved with simulation feedback and reuse them when acquiring new tasks.
SEAR~\cite{wang2026sear} evaluates such systems with budget-matched restarts and a no-library baseline to separate learning from resampling.
These methods demonstrate the potential of coding agents for autonomous improvement and skill accumulation.
However, they mainly focus on failure repair, policy improvement, or system evolution.
EvoMisty further focuses on performance optimization after task completion.
It also distills reusable functions from final solutions into a skill library, allowing future tasks to benefit from previous improvements.

\subsection{Agentic Code Performance Optimization}

Beyond functional correctness, coding agents have also been used for code performance optimization~\cite{shetty2026gso,he2025swe,ma2025swe,garg2026perfbench,cui2026perfopt,chen2026performance,opu2026agentic,novikov2025alphaevolve}. Peng et al.~\cite{peng2026agents} found that agent-authored performance optimization pull requests report explicit performance validation less often than human-authored ones. PerfAgent~\cite{deng2026perfagent} uses profiler feedback and test results to identify performance bottlenecks and guide iterative code optimization. However, robot task performance involves more than code execution efficiency. Metrics such as recognition accuracy, interaction latency~\cite{abbo2026multi,studerus2026framework}, and motion time require task-level measurements. These metrics cannot be fully evaluated by software profilers alone. EvoMisty uses data hooks and measurement hooks to collect execution data and performance evidence. It combines offline replay and physical robot evaluation to iteratively optimize programs based on user-specified metrics.

\section{Method}
\label{sec:method}

As shown in Fig.~\ref{fig:overview}, EvoMisty separates device-level capability construction from task-level execution. \textbf{Build} runs once per device and produces a fixed Atomic Capability Library shared across tasks. Each task is first completed by \textbf{Solve}; the user then decides whether to run \textbf{Optimize}, which improves the solution on user-specified metrics, and \textbf{Evolve}, which distills the final solution into the Skill Library.

\subsection{Build: Atomic Capability Library}
\label{sec:build}
In Build, EvoMisty guides the coding agent to read official device documentation, identify candidate capabilities, and construct a control program for verification and subsequent use. Since documented functions may behave differently on physical hardware, only capabilities verified on the robot are added to the Atomic Capability Library.
\paragraph{Physical Verification.}
For each candidate, the agent designs a test with specified inputs, expected outcomes, and pass conditions. It runs the test on the physical robot and evaluates the results using device states, execution logs, and other objective evidence when possible; otherwise, it requests human feedback. When a test fails, the agent iteratively repairs the control program and revalidates the affected capabilities; those that still fail after a fixed number of attempts are discarded.
\paragraph{Capability Encapsulation.}
The agent determines the actual functions and parameters of verified capabilities and encapsulates them as independently callable modules with usage guides. These modules form the Atomic Capability Library. New hardware can be integrated through the same process, without manually prepared device primitives.

\subsection{Solve: Task Completion}
\label{sec:solve}

Solve generates a task program using the Atomic Capability Library and the Skill Library, and verifies task completion on the physical robot. For each task, the agent divides the success criteria into verifiable conditions, which are confirmed by the user. Conditions based on execution logs, device responses, or events are checked by agent-generated programs. Conditions that require observing or listening to the robot are checked by humans.

The agent consults both libraries and prioritizes existing skills when writing the task program. After each run, EvoMisty first checks for execution errors and then runs the automatic checks. Human checks are performed only after all automatic checks pass. If any check fails, the agent revises the program based on the feedback and runs it again. A task is complete only when the program follows the required execution flow and satisfies all success criteria. Further performance improvements are handled separately by Optimize.

\subsection{Optimize: Metric-Guided Optimization}
\label{sec:optimize}

Solve ensures that a task executes successfully, but completing a task does not guarantee that it is performed well. If the user chooses to run Optimize, it improves the completed program with respect to user-specified metrics while preserving its original success criteria. Unlike conventional code optimization, the metrics that matter for robot tasks, such as recognition accuracy, interaction latency, and motion time, cannot be measured by software profilers alone. EvoMisty instead measures and improves performance using evidence collected from task execution.

\paragraph{Data Collection and Measurement.}
Based on the target metrics, EvoMisty guides the agent to instrument the program with two types of hooks. \emph{Data hooks} record raw inputs, such as audio and video, together with program outputs and robot feedback, which support later analysis and replay. \emph{Measurement hooks} record objective evidence, such as event timestamps and execution results. Performance metrics are computed from this evidence rather than estimated by the agent.

EvoMisty first establishes a performance baseline by executing the program on the physical robot. For components that can be evaluated offline, it replays the recorded data so that program changes are compared under identical inputs. Closed-loop tasks, such as line following, instead require complete physical runs, because the robot's actions affect its subsequent observations; in these cases, replay is used only to analyze problems.

\paragraph{Iterative Optimization and Selection.}
In each iteration of the optimization loop, the agent identifies a performance bottleneck from the measurements and makes a single change to the program.
The resulting candidate is evaluated on the target metrics and checked against the original success criteria.For valid candidates, EvoMisty maintains a Pareto
frontier of non-dominated solutions based on the measured metrics.
Optimization stops when the predefined budget is exhausted or when no further improvement is observed over several consecutive iterations.

\subsection{Evolve: Skill Distillation}
\label{sec:evolve}

Evolve distills reusable functions from a final task solution into the Skill Library, so that future tasks can build on them instead of starting from the atomic capabilities alone. When the solution has been optimized, its performance improvements carry over to future tasks through the distilled skills. The Skill Library is updated only when the user chooses to run Evolve.

Distillation proceeds in three steps. First, EvoMisty guides the agent to \emph{extract} candidate skills: the agent identifies independently callable functions in the final program, removes their task-specific dependencies, and encapsulates each one as a candidate. Second, each candidate is \emph{verified}: EvoMisty checks that it remains consistent with the final program and confirms that it is callable through simulated dependency checks. Third, each verified candidate is \emph{compared} with the existing skills in the library. A candidate that provides new functionality is added as a new skill; one that improves on an existing skill is saved as a new version, with previous versions preserved; and a duplicate is discarded. Accepted skills are stored with interface descriptions and usage guides, which support their reuse and composition in future tasks.

\section{Experiments}
\label{sec:experiments}
We evaluate EvoMisty through four research questions:
\begin{itemize}
    \item \textbf{RQ1: Capability Construction.} \textit{Given only the official documentation, can the agent autonomously build device capabilities verified on the physical robot?}

    \item \textbf{RQ2: Task Completion.} \textit{Can EvoMisty complete the AutoMisty benchmark tasks, as well as tasks that AutoMisty cannot complete?}

    \item \textbf{RQ3: Metric-Guided Optimization.} \textit{Can EvoMisty improve a completed program on user-specified metrics?}

    \item \textbf{RQ4: Skill Reuse.} \textit{Can skills distilled from one task be reused by, and benefit, other tasks?}
\end{itemize}
\paragraph{Experimental Setup.}
All experiments use a Misty~II social robot. To test whether the results depend on a particular agent, we run EvoMisty on two coding agents, Claude Code (Opus~5.5) and Codex (GPT-6 Astra), both with reasoning effort set to \texttt{high}. Our baseline is AutoMisty, a carefully hand-designed framework for Misty. Since the original AutoMisty uses GPT-4o, we replace its backbone with Opus~5.5 and rerun all tasks to rule out backbone differences. Every run starts in a fresh container with no memory and no access to previous trajectories; only the Atomic Capability Library and the Skill Library persist across runs. 
\label{sec:rq1}
\begin{table}[t]
\centering
\small
\setlength{\tabcolsep}{3.5pt}
\caption{Capability construction results.}
\label{tab:rq1}
\begin{tabular}{@{}lccccc@{}}
\toprule
Method & Extracted & Excluded & Tested & Failed & Verified \\
\midrule
AutoMisty~\cite{wang2025automisty} & --  & -- & --  & -- & 12  \\
Claude Code                & 175 & 63 & 112 & 3  & 109 \\
Codex                      & 175 & 63 & 112 & 5  & 107 \\
\bottomrule
\end{tabular}

\vspace{2pt}

\end{table}

\subsection{RQ1: Capability Construction}
\begin{figure*}[t]
  \centering
  \includegraphics[width=\linewidth]{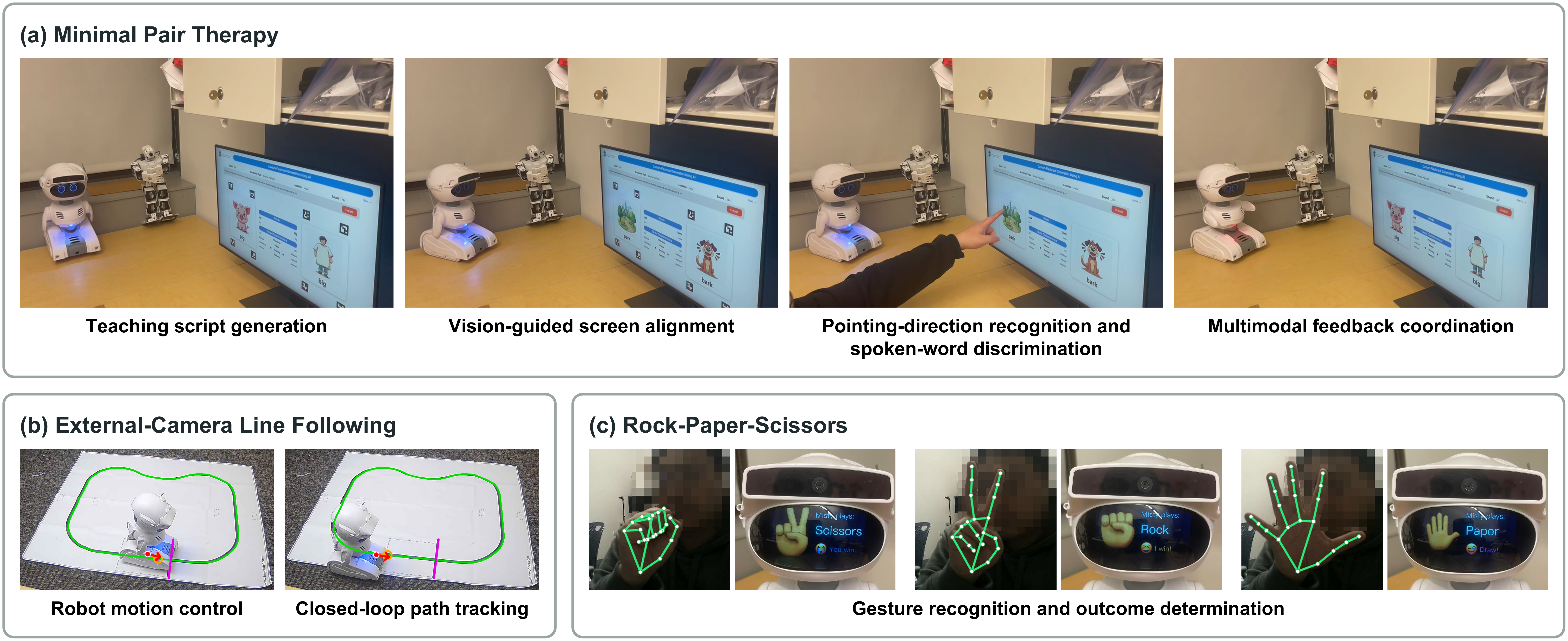}
  \caption{Representative new tasks. Each panel shows the main steps
of one task as executed on the Misty robot.}
  \label{fig:tasks}
\end{figure*}

EvoMisty takes only the official Misty documentation as input, autonomously builds device capabilities, and verifies each of them on the physical robot. Throughout the process, a human only answers binary questions when the system asks, such as ``Did the robot move forward?'' or ``Did it speak?''. As shown in Table~\ref{tab:rq1}, EvoMisty extracts 175 candidate capabilities from the documentation and autonomously excludes 63 of them from testing because the robot lacks the required hardware; this decision agrees with our manual review. The remaining 112 are tested on the robot, of which Claude Code verifies 109 and Codex verifies 107. Our manual review confirms that 110 of the 112 are actually usable, so the two agents cover 109 and 107 of them, respectively.

The failures fall into three categories. First, \emph{overly strict judgment}: onboard speech recognition works in practice, but its return values differ from the documentation, and both agents therefore mark it as failed. Second, \emph{no observable verification condition}: for clearing on-screen error messages and for hazard detection events, the system cannot construct a triggering condition, so neither agent can verify them. Third, \emph{unstable capabilities}: driving by left and right track speeds and world-state events fail only with Codex; our manual review finds that both behave intermittently and happened to malfunction during the Codex run. Overall, verification is conservative: no unusable capability is accepted as usable.

EvoMisty also achieves substantially broader coverage of the robot's native capabilities than AutoMisty's hand-designed framework. AutoMisty~\cite{wang2025automisty} directly uses 12 native capabilities, on top of which its authors manually wrapped 136 capabilities for motions and expressions. As Table~\ref{tab:rq1} shows, both agents verify all 12 of these, and EvoMisty covers about nine times as many native capabilities. The same process extends to external hardware: from the device documentation, EvoMisty builds control code for a gamepad, a standalone keypad, and an external camera, and confirms on the physical robot that each works. 

In summary, given only the official documentation and a small amount of binary feedback, EvoMisty autonomously builds a set of device capabilities that is broader than the hand-designed one and extends to new hardware, with largely consistent results across the two agents.

\subsection{RQ2: Task Completion}
\begin{table*}[t]
\centering
\small
\renewcommand{\arraystretch}{1.15}
\setlength{\tabcolsep}{5pt}
\caption{Optimization results across tasks and metrics.}
\label{tab:rq3}
\begin{tabular}{|l|l|c|c|c|c|}
\hline
 & & \multicolumn{2}{c|}{\textbf{Claude Code}} & \multicolumn{2}{c|}{\textbf{Codex}} \\
\cline{3-6}
\textbf{Task} & \textbf{Metric} & Before & After & Before & After \\
\hline
Naughty-Child Dialogue & Response latency (s) $\downarrow$ & $7.41 \pm 0.70$ & $3.54 \pm 0.47$ & $7.18 \pm 1.03$ & $2.06 \pm 0.25$ \\
\cline{2-6}
 & CER $\downarrow$ & 0.053 & 0.000 & 0.037 & 0.000 \\
\hline
MistyPilot Emotional Dialogue & Response latency (s) $\downarrow$ & $5.59 \pm 1.63$ & $4.51 \pm 0.56$ & $6.62 \pm 0.60$ & $4.47 \pm 0.46$ \\
\hline
Rock-Paper-Scissors & Gesture recognition accuracy $\uparrow$ & 0.845 & 0.987 & 0.273 & 0.874 \\
\hline
External-Camera Line Following & Lap time (s) $\downarrow$ & 32.0 & $20.3 \pm 6.2$ & 733 & $339 \pm 11$ \\
\hline
Minimal Pair Therapy & Pointing-direction accuracy $\uparrow$ & 0.833 & 0.917 & 0.667 & 0.750 \\
\cline{2-6}
 & Spoken-word discrimination accuracy $\uparrow$ & 0.667 & 0.917 & 0.500 & 0.917 \\
\hline
\end{tabular}

\vspace{2pt}
\parbox{\textwidth}{\footnotesize Time metrics are reported as mean $\pm$ standard deviation. $\downarrow$: lower is better; $\uparrow$: higher is better.}
\end{table*}

We evaluate on two sets of tasks. The first is the full AutoMisty~\cite{wang2025automisty} benchmark of 28 tasks, which span seven categories: body motion, expression and speech output, multimodal performance, touch interaction, voice commands and question answering, visual understanding, and spoken dialogue. The second is a set of 7 new tasks that test whether a system can go beyond what a hand-designed framework anticipates. They are designed around four kinds of requirements that the AutoMisty framework does not support: native capabilities that it does not wrap, external hardware, external resources, and perception algorithms that the robot does not natively provide. The new tasks fall into two groups. Two are small extensions of benchmark tasks: Original-Voice Pikachu extends AutoMisty's Pikachu task by retrieving an original Pikachu voice clip from the web and playing it while displaying a Pikachu image. Face-Tracking Recording extends AutoMisty's recording task by rotating the robot's base to follow a detected face during recording. The other five are substantially harder than any benchmark task. MistyPilot Emotional Dialogue ports the emotional expression mechanism of MistyPilot~\cite{wang2026mistypilot} from its source code and uses the keys of an external keypad to control the dialogue. Gamepad Teleoperation integrates a gamepad to teleoperate the robot like a remote-controlled car. Rock-Paper-Scissors (Fig.~\ref{fig:tasks}c) builds its own gesture recognition: it recognizes the user's gesture, displays the robot's move on its screen, and determines the outcome. External-Camera Line Following (Fig.~\ref{fig:tasks}b) localizes the robot with a network camera: it detects the track and the robot's position in the external camera view and uses them for closed-loop steering correction over one lap of a loop track. Minimal Pair Therapy (Fig.~\ref{fig:tasks}a) is adapted from a clinical speech-language intervention for children with speech sound disorders~\cite{weiner1981treatment,barlow2002minimal}: the robot first generates the teaching script, then uses vision to locate the screen and find a suitable viewing angle, judges which card the child points at and which word of a pair the child says, and finally coordinates multimodal feedback.For each task, we evaluate completion at the workflow level. A task is considered complete when its program
executes the required end-to-end workflow on the physical robot, including the necessary perception, decision-making, and action stages. Completion does not require meeting
quantitative performance thresholds, such as recognition accuracy, response latency, or execution time, which are evaluated separately in RQ3. EvoMisty and AutoMisty are
evaluated using the same completion criteria for each task.

On the 28 benchmark tasks, EvoMisty completes all of them with both agents, matching AutoMisty. AutoMisty relies on 136 manually wrapped capabilities and a hand-designed execution framework, whereas EvoMisty uses only the device capabilities it builds autonomously, which shows that these capabilities can replace manual wrapping and design.

The difference appears on the 7 new tasks: EvoMisty completes all of them with both agents, while AutoMisty,Even after upgrading AutoMisty's backbone from GPT-4o
to the more capable Opus 5.5, it cannot complete any of the seven new tasks. This suggests that its limitations stem from its fixed framework rather than simply from the capability of the backbone model. The reason is that its framework wraps only 12 native capabilities; it can neither drive the base nor upload files, and it has no way to connect external devices. Even the two extension tasks are out of reach, the former for lack of an capabilities to upload audio and images and the latter for lack of an capabilities to drive the base. The limitation therefore comes from the fixed framework, not from the model.

\subsection{RQ3: Metric-Guided Optimization}

\begin{table*}[t]
\centering
\small
\renewcommand{\arraystretch}{1.15}
\setlength{\tabcolsep}{4pt}
\caption{Skill reuse results on the four dialogue tasks.}
\label{tab:rq4-transfer}
\resizebox{\textwidth}{!}{%
\begin{tabular}{|l|l|c|c|c|c|c|}
\hline
 & & & \multicolumn{2}{c|}{Claude Code} & \multicolumn{2}{c|}{Codex} \\
\cline{4-7}
Task & Metric & AutoMisty & Before & After & Before & After \\
\hline
Naughty-Child Dialogue & Response latency (s) & $4.70 \pm 2.24$ & $7.41 \pm 0.70$ & $3.54 \pm 0.47$ & $7.18 \pm 1.03$ & $2.06 \pm 0.25$ \\
\cline{2-7}
(optimized) & CER & 0.077 & 0.053 & 0.000 & 0.037 & 0.000 \\
\hline
Math Tutoring & Response latency (s) & $4.54 \pm 1.10$ & $7.29 \pm 1.17$ & $4.26 \pm 0.33$ & $9.49 \pm 2.14$ & $3.12 \pm 0.38$ \\
\cline{2-7}
(skill reuse) & CER & 0.306 & 0.061 & 0.019 & 0.024 & 0.019 \\
\hline
Keyword Story & Response latency (s) & $6.10 \pm 3.02$ & $16.58 \pm 3.47$ & $4.63 \pm 0.69$ & $10.92 \pm 1.11$ & $6.61 \pm 0.50$ \\
\cline{2-7}
(skill reuse) & CER & 0.433 & 0.200 & 0.117 & 0.400 & 0.000 \\
\hline
Chinese-to-English Translation & Response latency (s) & $3.37 \pm 0.37$ & $5.86 \pm 0.76$ & $2.97 \pm 0.36$ & $5.40 \pm 1.09$ & $3.61 \pm 0.30$ \\
\cline{2-7}
(skill reuse) & CER & 0.059 & 0.032 & 0.018 & 0.136 & 0.017 \\
\hline
\end{tabular}%
}

\vspace{2pt}
\parbox{\textwidth}{\footnotesize
\emph{Optimized}: Before/After denote before/after optimization.
\emph{Skill reuse}: Before/After denote without/with distilled skills, without further optimization.
Latency is reported in seconds (mean $\pm$ std).
}
\end{table*}

We evaluate Optimize on 7 metrics across 5 tasks (Table~\ref{tab:rq3}): one benchmark task and four new tasks, covering speech interaction, visual perception, and motion control. We select tasks that have a quantifiable performance metric beyond binary success.Among the four benchmark dialogue tasks, which share the same pipeline, we optimize only Naughty-Child Dialogue, in which the robot plays a naughty child and engages in a multi-turn conversation with a person, and reserve the other three tasks for the skill reuse experiment in Sec.~\ref{sec:rq4-similar}. For each task, the user specifies the metrics in natural language. All tasks use the same optimization procedure, with no design specific to any task or metric.

Each metric corresponds to an optimization target that the user specifies in the task definition, and all measurements follow three principles. First, the values in Table~\ref{tab:rq3} are measured independently by us. Second, the test data is never used during optimization. Third, the programs before and after optimization are measured with the same protocol.

\noindent The metrics are defined as follows:
\begin{itemize}
  \item \textbf{Response latency}: the time from the end of the user's utterance to the moment the robot starts speaking, measured from audio recordings over 10 turns for each of the two dialogue tasks.
  \item \textbf{Character error rate (CER)}: compares the system's transcription of the user's speech with a human-annotated reference, character by character.
  \item \textbf{Gesture recognition accuracy}: computed per frame on 10 videos with 146 frames in total, as the average of the accuracies for rock, paper, and scissors; frames in which no hand is detected or no decision is made count as errors.
  \item \textbf{Lap time}: the time from when the robot crosses the start line until it crosses it again after one full lap along the track, timed manually, with no shortcuts or departures from the track allowed; we time 1 lap before optimization and 3 laps after.
  \item \textbf{Pointing-direction accuracy}: whether the system correctly decides if the user points to the left or the right card, over 12 trials with manually recorded ground truth.
  \item \textbf{Spoken-word discrimination accuracy}: whether the system correctly decides which word of the pair the user says, over 12 trials with manually recorded ground truth. 
\end{itemize}

With both agents, all 7 metrics improve. Among the time metrics, the response latency of Naughty-Child Dialogue drops by 52\% and 71\% (Claude Code and Codex, respectively; the same order is used below), that of MistyPilot Emotional Dialogue by 19\% and 32\%, and the lap time of line following by 37\% and 54\%. Among the recognition metrics, gesture recognition accuracy rises from 0.845 and 0.273 to 0.987 and 0.874, which reduces the error rate by 92\% and 83\%; CER falls to 0 for both agents; and for Minimal Pair Therapy, spoken-word discrimination accuracy rises from 0.667 and 0.500 to 0.917 for both agents, while pointing-direction accuracy also improves (from 0.833 to 0.917 and from 0.667 to 0.750).

The initial programs differ widely in performance because the two agents choose different implementations: the initial lap times are 32.0\,s and 733\,s, and the initial gesture recognition accuracies are 0.845 and 0.273. All of these initial programs already satisfy the success criteria of their tasks and differ only in performance. Optimization improves the program regardless of where it starts, but the level it reaches still depends on the initial implementation: the line-following program that starts at 733\,s reaches 339\,s, whereas the one that starts at 32.0\,s reaches 20.3\,s.

In summary, the answer to RQ3 is yes: the same procedure improves all 7 user-specified metrics across 5 tasks with both agents. Optimize delivers improvement relative to the starting point, and the final level also depends on the implementation that the agent initially chooses.

\subsection{RQ4: Skill Reuse}
We examine Evolve in two settings:

\begin{itemize}
    \item \textbf{Similar tasks.} \textit{Can the optimization gains of one task transfer to other tasks through skills and bring measurable benefits?}

    \item \textbf{Different tasks.} \textit{Can the skills distilled from several tasks be composed in a new task?} 
\end{itemize}

\subsubsection{Skill Reuse Across Similar Tasks}
\label{sec:rq4-similar}

The AutoMisty benchmark contains four dialogue tasks that follow the same pipeline: recognize the user's speech, generate a reply, synthesize speech, and play it on the robot. They differ only in how the reply is generated. We run Optimize \textbf{only} on Naughty-Child Dialogue, targeting response latency and CER, and then run Evolve to distill its final program into skills. For each of the other three tasks (Math Tutoring, Keyword Story, and Chinese-to-English Translation), the agent solves the task twice: once with an \textbf{empty Skill Library} and once \textbf{with these skills}. Both runs start in a fresh container and go through Solve only, without Optimize, so the only difference is whether the skills are available.

With the skills, both metrics improve on all three tasks with both agents, in all 12 comparisons (Table~\ref{tab:rq4-transfer}). Response latency drops by 42\% and 67\% on Math Tutoring (Claude Code and Codex, respectively; the same order is used below), by 72\% and 39\% on Keyword Story, and by 49\% and 33\% on Chinese-to-English Translation, an average reduction of 50\% over the six comparisons. CER drops by 69\% and 21\% on Math Tutoring, by 42\% and 100\% on Keyword Story, and by 44\% and 88\% on Chinese-to-English Translation, an average reduction of 60\%. Because the two runs differ only in whether the skills are available, these improvements can be attributed to skill reuse; none of the three tasks is optimized itself.

The dialogue pipeline of AutoMisty was manually tuned for latency. Without skills, both agents are slower than AutoMisty on all four tasks. After optimization or skill reuse, Claude Code is faster than AutoMisty on all four tasks, and Codex is faster on two of them. CER is already lower than that of AutoMisty in most cases beforehand, and it is lower on all four tasks afterwards. 

These results demonstrate that making distilled skills available improves task performance without further optimization.

\subsubsection{Skill Composition Across Different Tasks}
\label{sec:rq4-different}

We let both agents complete, in sequence, the five harder new tasks introduced in Sec.4.2. MistyPilot Emotional Dialogue, Gamepad Teleoperation, Rock-Paper-Scissors, External-Camera Line Following, and Minimal Pair Therapy. Each task runs in a fresh container, and only the Skill Library carries over: after each of the first four tasks, EvoMisty distills its reusable functions into skills, so that when the last task starts, the Skill Libraries of Claude Code and Codex contain 21 and 18 skills, respectively. We then inspect every skill call in the final program of Minimal Pair Therapy and trace each skill back to the task it was distilled from. This task is a demanding test of composition, as it simultaneously requires image capture, pointing-direction recognition, emotional speech, spoken-word discrimination, and rotating the base to face the screen.

Most of the capabilities that Minimal Pair Therapy requires are already available to both agents from earlier tasks. The Claude Code program draws on all four preceding tasks: to capture the cards and the pointing finger, it uses the camera video stream from Gamepad Teleoperation; to speak each line with its specified emotion, it uses speech style setting from MistyPilot Emotional Dialogue; to accompany speech with chest-light and arm motions, it uses emotional motion from Rock-Paper-Scissors; and to track the heading while the base rotates, it uses heading readout from External-Camera Line Following. The Codex program draws on two of them: to speak with emotion, it reuses emotional speech playback from MistyPilot Emotional Dialogue; to turn the base toward the screen, it reuses heading-feedback driving from External-Camera Line Following; and to tell which word of the pair the user says, it rewrites the speech recognition skill from MistyPilot Emotional Dialogue. What both agents still write from scratch is only the logic specific to this task, such as card recognition, pointing-direction recognition, and the teaching protocol.

In summary, the answer to RQ4 is yes. The Skill Library stores independently callable modules rather than complete task programs. These modules carry the optimization gains of one task over to similar tasks, and they can be reused directly or rewritten and composed with new task logic in different tasks.

\section{Limitations}
\label{sec:limitations}

Our experiments focus on the Misty~II social robot. Although we evaluate
two coding agents and integrate several external devices, the
applicability of EvoMisty to other robot platforms remains to be
validated. Moreover, capability verification and task completion checks
still rely on human observation and feedback when device states,
execution logs, or events do not provide sufficient evidence.
Verification coverage is also constrained by the test conditions that
can be constructed. Finally, the unified optimization workflow can
improve task performance, but the final performance still depends on
the initial implementation chosen by the coding agent, and substantial
differences between implementations may remain after optimization.

\section{Conclusion}
\label{sec:conclusion}
Existing LLM-based methods for social robots mostly rely on hand-built
capabilities and stop once a task is completed. EvoMisty removes both
limitations: coding agents build the capabilities themselves and keep
improving completed tasks on user-specified metrics. We validate three
contributions on a physical Misty~II with Claude Code and Codex.
First, from the official documentation and a small amount of binary
human feedback, the two agents build and verify 109 and 107 atomic
capabilities, respectively, covering about nine times as many native
APIs as the hand-designed AutoMisty. With these capabilities, EvoMisty
completes all 28 benchmark tasks and 7 new tasks that AutoMisty cannot
complete with the same backbone.
Second, a single procedure, driven by evidence collected during
execution and with no design specific to any task or metric, improves
all 7 user-specified metrics across 5 tasks spanning speech, vision,
and motion control.
Third, skills distilled from one optimized dialogue task reduce
response latency by 50\% and CER by 60\% on average in three other
dialogue tasks that are never optimized. In Minimal Pair Therapy, the
agents compose skills from up to four different tasks and write only
the task-specific logic.

\clearpage
\section{LLM Usage Statement}
\label{sec:llm}
LLM-based coding agents are the subject of this work: Claude Code
(Opus~5.5) and Codex (GPT-6 Astra) were used to generate, test, and
optimize robot programs in our experiments. GPT-6 Astra was used to
generate some of the robot illustrations in our figures. Claude Fable~5.2
and GPT-6 Astra were used to check grammar and improve the clarity of the
writing. The authors reviewed and verified all LLM-assisted content and
take full responsibility for the paper.

{
    \small
    \bibliographystyle{ieeenat_fullname}
    \bibliography{main}
}

\clearpage
\onecolumn
\setcounter{page}{1}
\setcounter{section}{0}
\renewcommand{\theHsection}{supp.\arabic{section}}
\input{sec/X_suppl}

\end{document}